\documentclass{article}
\usepackage[utf8]{inputenc}
\usepackage{DejaVuSansMono}
\usepackage{authblk}
\usepackage{setspace}
\usepackage[margin=1.25in]{geometry}
\usepackage{graphicx}
\graphicspath{ {./figures/} }
\usepackage{subcaption}
\usepackage{amsmath}
\usepackage{amssymb}
\usepackage{mathtools, cuted}
\usepackage{fancyvrb}
\usepackage{tcolorbox}

\usepackage{bbm}
\usepackage{courier}
\usepackage{hyperref}  

\usepackage{cprotect}
\usepackage{listings}

\usepackage{algorithm}
\usepackage[noend]{algpseudocode}  

\algrenewcommand\algorithmicrequire{\textbf{Defining:}}

\usepackage{tcolorbox}

\usepackage{xcolor}
\definecolor{codegreen}{rgb}{0,0.6,0}
\definecolor{codegray}{rgb}{0.5,0.5,0.5}
\definecolor{codepurple}{rgb}{0.58,0,0.82}
\definecolor{backcolour}{rgb}{0.95,0.95,0.92}
\lstdefinestyle{mystyle}{
  backgroundcolor=\color{backcolour}, commentstyle=\color{codegreen},
  keywordstyle=\color{magenta},
  numberstyle=\tiny\color{codegray},
  stringstyle=\color{codepurple},
  basicstyle=\ttfamily\footnotesize,
  breakatwhitespace=false,         
  breaklines=true,  
  breakindent=80pt,
  captionpos=b,                    
  keepspaces=true,                  
  numbersep=5pt,                  
  showspaces=false,                
  showstringspaces=false,
  showtabs=false,                  
  tabsize=2,
  xleftmargin=0pt,   
  xrightmargin=0pt   
}
\usepackage[style=ieee, 
citestyle=numeric-comp,backref=true,hyperref=true, indexing=true,
sorting=nyt]{biblatex}
\title{SLEM: Machine Learning for Path Modeling and Causal Inference with Super Learner Equation Modeling}

\author[]{Matthew J. Vowels\textsuperscript{1}}

\affil[]{\textsuperscript{1} Institute of Psychology, \\ 
University of Lausanne, Lausanne, Switzerland.}

\date{}

\begin{document}

\maketitle


\begin{abstract}
Causal inference is a crucial goal of science, enabling researchers to arrive at meaningful conclusions regarding the predictions of hypothetical interventions using observational data. Path models, Structural Equation Models (SEMs), and, more generally, Directed Acyclic Graphs (DAGs), provide a means to unambiguously specify assumptions regarding the causal structure underlying a phenomenon. Unlike DAGs, which make very few assumptions about the functional and parametric form, SEM assumes linearity. This can result in functional misspecification which prevents researchers from undertaking reliable effect size estimation. In contrast, we propose Super Learner Equation Modeling, a path modeling technique integrating machine learning Super Learner ensembles. We empirically demonstrate its ability to provide consistent and unbiased estimates of causal effects, its competitive performance for linear models when compared with SEM, and highlight its superiority over SEM when dealing with non-linear relationships. We provide open-source code, and a tutorial notebook with example usage, accentuating the easy-to-use nature of the method.
\end{abstract}

\section{Introduction}
\label{sec:intro}
Imagine you wish to estimate multiple causal effects from observational data. Despite the challenging nature of the task, it nonetheless represents one of the core goals of science and causality \parencite{Pearl2018TBOW, Pearl2012, Vowels2021, Vowels2021DAGs}. Indeed, in the absence of experimental data, we must do everything we can to ensure the causal relevance of our statistical inferences. Otherwise, our estimates cannot be tied to their associated theories \parencite{Vowels2021, Scheel2020} and the estimates effectively represent arbitrary functions of the observed data, subject to ambiguous, pseudo-causal interpretations \parencite{Grosz2020, Rohrer2018, Hernan2018}. To this end, researchers in the domains of psychology and social science have begun to advocate for the increased adoption of causal Directed Acyclic Graphs (DAGs), which aid in (a) the clear, formal specification of a causal theory as a mathematical but nonetheless intuitive, visual object, (b) unbiased estimation of the target quantities of estimation (\textit{e.g.}, effect sizes) \parencite{Vowels2023_prespec, Rohrer2018}. 

Unfortunately, DAGs themselves only get us so far in terms of achieving unbiased estimation of causal effects. Whilst they help us deal with the specification of the \textit{structural} aspect of a model/theory (by representing an ordered causal process), and whilst they enable us to express a target quantity as a function of the observed distribution, they do not help with the estimation itself. Furthermore, they do not help us with the \textit{functional} specification of the relationships between variables. Indeed, one of the strengths of DAGs is that they are non-parametric, and make very limited assumptions about the underlying functional form. For instance, they tell us nothing about whether $Y$ is a linear function of just $X$ or whether it is a linear function of $X^2$. Thus, to arrive at meaningful effect size estimates, the choice of the associated estimation technique and its associated level of functional adaptability must be made. One option that researchers have is to use structural equation modeling or linear/logistic regression for the estimation of the associated effects, which make the assumption that the relationships are linear/linear in the logit space. The problem with this is that such an assumption of linearity can lead to a similar level of biased estimation \parencite{Vowels2021, vanderLaan2011, vanderLaan2014} to that which results from structural misspecification. Such misspecification thereby undermines the otherwise advantageously function-agnostic nature of the DAG. 

In order to avoid making such unnecessary assumptions about the functional form underlying the causal relationships represented in the DAG, we recommend the use of machine learning. Specifically we propose Super Learner Equation Modeling (SLEM), an integrated framework for machine learning based causal inference with DAGs. Within this SLEM framework, we present the DAG Learner estimator object in the form of easy-to-use, open-source Python code,  including simulations and a tutorial-style notebook.\footnote{The code can be found in supplementary material or at \url{https://github.com/matthewvowels1/SLEM}.} Essentially, for any user-specified DAG, and according to the SLEM framework, the DAG Learner implements a set of general machine learning models which are used to estimate, in a data-driven but causally-constrained manner, all associated path coefficients. Furthermore, our framework allows for the estimation of the effect of (optionally multiple, simultaneous) user-specified interventions, thereby facilitating a general, and easy-to-use tool for non-parametric causal inference. In our view, the availability of such a tool is overdue - whilst the required techniques have existed for some time, they have never been combined in such a way.

The paper is structured as follows: First, we briefly discuss some background theory relating to DAGs, causality, and machine learning. Secondly, we describe the proposed framework, beginning with the specification of the DAG, as well as the chosen machine learning method. Then, we provide some worked examples and key simulations. Finally we discuss the associated limitations of the method in relation to some existing alternative approaches, and close with a summary. Note that whilst the success of the propose method rests on existing work in the domains of causality and machine learning, to the best of our knowledge we are the first to bring the two together in such an integrated, easy-to-use methodology.

\section{Background}
\label{sec:background}

In this section, we review pertinent technical concepts related to Directed Acyclic Graphs (DAGs), causality, and machine learning. For a comprehensive understanding, readers can refer to various established resources \parencite{Koller2009, Pearl2009, Peters2017, bishop, murphy2}. In terms of notation, we use symbols $X$ or $A, B, C$ etc. to symbolize random variables, with bold symbols like $\mathbf{X}$ denoting a set of variables, and lower-case letters $x$ or $\mathbf{x}$ indicating specific values for variables, for example $X=x$ or  $\mathbf{X}=\mathbf{x}$.

\subsection{Causal Directed Acyclic Graphs}
\label{sec:dags}
Briefly, a causal DAG (hereafter, simply referred to as a DAG) is a graphical representation of a set of causal-effect links in the form of a set of nodes/variables and directed arrows. Their acyclicity prohibits the presence of closed loops or cycles. An example of a DAG is $X \rightarrow Y$, which implies that, according to our theory or model of a phenomenon, $X$ causes $Y$. One can also say that $X$ is a causal parent of $Y$, and that $Y$ is a child of $X$. 

Apart from the general nature of DAGs (avoiding, as they do, any necessary specification of functional or distributional form), one of the benefits of their usage is the associated causal $do$-calculus \parencite{Pearl2009}, which enables researchers to express causal quantities in terms of the observed, joint distribution. For example, in contrast to the usual $P(Y=y|X=x)$, which is a conditional probability statement meaning the `probability of $Y$ being equal to $y$ given that $X$ is equal to $x$, $P(Y=y|do(X=x))$ is a hypothetical causal statement, meaning the `probability of $Y$ being equal to $y$, given the \textit{intervention setting} $X$ to the value $x$. The key difference here is that if one intervenes on $X$ in such a way, one removes any dependence that $X$ otherwise had on its parents, and also no longer necessarily takes on the originally observed values. The reason this statement refers to a somewhat hypothetical quantity is because, in the absence of experimental data, such an intervention is not possible with observational data (and if it were, one would therefore have access to experimental conditions).

In the simple graph $X \rightarrow Y$, it so happens that $P(Y=y|X=x) = P(Y=y|do(X=x))$. In other words, the observed quantity is equivalent to the interventional quantity. However, introduce a third \textit{confounding} variable $C$ which is a parent of both $X$ and $Y$, and now $P(Y=y|X=x) \neq P(Y=y|do(X=x))$. In this case, one must adjust/control for $C$, and by doing so, $P(Y=y|X=x, C) = P(Y=y|do(X=x), C)$, thus making an otherwise hypothetical/theoretical quantity estimable from observational data. More generally, \textit{do}-calculus provides a framework for (where possible) making $P(Y=y|\mathbf{S}, X) = P(Y=y|\mathbf{S}, do(X=x))$, where $\mathbf{S}$ is a set of necessary adjustment variables. In this way, under a set of strong assumptions (such as the assumption that the assumed DAG model is a sufficient representation of the underlying causal reality), causal inference is possible even with observational data.

If one wishes to use \textit{do}-calculus to estimate the Average Treatment Effect, for example, one might be interested in estimating the difference of two expected values. For example, $\mathbb{E}[Y=1|do(T=1)) - \mathbb{E}[Y=1|do(T=0))$ could represent the difference in the likelihood of recovery ($Y=1$) under two treatment conditions ($T=1$ and $T=0$). Note that under randomized, experimental conditions, this quantity is estimable by equivalently computing the difference in the average outcomes for those in the treatment ($T=1$) versus the control ($T=0$) groups.

In short, the key advantage of DAGs is that they provide a formal representation of our assumptions about the causal structure underlying the phenomenon of interest. Together with $do$-calculus, DAGs provide the means to translate a hypothetical quantity (written in terms of \textit{do} notation) into estimable quantities. This is not possible in all cases (notably when there exist certain unobserved third variables), but the common alternative - which often involves vague verbal justifications for the inclusion of certain control variables and an absence of formalisation - is generally less transparent and more prone to subjective variability.

\subsection{Structural Causal Models}
\label{sec:scm}
DAGs can be translated into a set of structural equations, and this set is known as a Structural Causal Model (SCM). It is a generalisation of the well-known Structural Equation Model which assumes that the modeled relationships between variables are fundamentally linear. Consider again the DAG $X \rightarrow Y$. This can be represented as a simple SCM:

\begin{equation}
\begin{split}
    X := f_X(U_X)\\
    Y := f_Y(X, U_Y),
\end{split}
\end{equation}

where the $:=$ symbols clarify the causally asymmetric nature of the relationship ($X$ causes $Y$, not the other way around), $f$ is a general function relating the right hand side of the equation to the left, and $U$ represents some additional stochasticity, such that $Y$ is not a deterministic function of its parent $X$, and that $X$ also randomly varies according to some underlying random variation. Note that SEMs posit that all internal variables result from a linear combination of others, and in such as case the SCM above be expressed as $Y := \beta_{XY}X + U_Y$. Here, the $\beta$ is a structural parameter (or path coefficient) which represents the causal effect size of $X$ on $Y$.

Introducing a partial mediator $M$ to the graph above, and the associated SCM becomes:

\begin{equation}
\begin{split}
   X := f_X(U_X)\\
   M:= f_M(X, U_M)\\
   Y := f_Y(X, M, U_Y). 
\end{split}
\end{equation}

Note that, if the effect of $X$ on $Y$ were \textit{fully} mediated by $M$, $X$ would not appear in the equation for $Y$. Note also, that according to $do$-calculus, an intervention, for example $do(M=2)$, would be represented as: 

\begin{equation}
\begin{split}
   X := f_X(U_X)\\
   M:= 2\\
   Y := f_Y(X, M=2, U_Y). 
\end{split}
\end{equation}

Note that the dependence that $M$ had on $X$ is removed, and the value is fixed to 2. This is known as a hard-intervention (soft interventions can be used to specify a distribution for the variable being intervened on). Similarly, such an intervention essentially severs the incoming causal influences, effectively removing the parents of the variable under intervention.

\subsection{Super Learning}
\label{sec:ml}
In the SCMs above, we used $f_Y$ to denote the function relating $Y$ to its causal parents in the DAG, and we mentioned that in the SEM framework, this function is either assumed to be linear, or the non-linearity must be specified \textit{a priori} \parencite{Umbach2017, Wen2010}.\footnote{The `product-indicator' class of SEM approaches have this requirement \parencite{Umbach2017, Marsh2006}.} In order to avoid making assumptions about linearity which can result in biased estimation \parencite{vanderLaan2014, Vowels2021}, we prefer to allow this function to be estimated from the data themselves. Machine learning provides an enormous set of algorithmic solutions for estimating such functions.\footnote{For a broad introduction to machine learning in the context of psychology and social science, readers are encouraged to review \textcite{Yarkoni2017}.}

For our purposes, we conceptualize machine learning algorithms as black boxes which take in a set of input predictors, and output a prediction for the specified outcome. Machine learning models are able to perform this task through a process known as training, whereby a given criterion is optimized (\textit{e.g.}, linear regression, which can be considered to be a machine learning algorithm, is optimized to minimize the squared error by solving an estimating equation) during what is known as a training process. It is generally recommended that the trained algorithm is then tested with respect to its ability to predict the outcome values for data it did not see during training. In the case of linear regression, the machine learning algorithm is constrained to find the parameters of a weighted linear sum of the predictor variables which minimizes the squared error for the predictions. For more complex algorithms, such as MultiLayer Perceptrons, which are a type of neural network, the algorithm is tasked with finding a suitable set of parameter values (which often number in the thousands, millions, or billions) which achieve the same thing as linear regression, but with much more flexibility/non-linearity. Each algorithm is biased towards a certain type of function class - linear regression is constrained to find a straight line solution, whereas decision trees, for example,  create a decision logic which results in the discontinuous, hard-segmentation of the input predictors according to their values.

Whilst a few specific algorithms have seen sporadic adoption in psychology and social science \parencite{Joel2020, Hilpert2023, Biggiogera2021, VowelsPartnerSupport}, particularly the random forest \parencite{breiman2001}, researchers in the domain of statistics, epidemiology, and machine learning have proposed the Super Learner \parencite{Polley2007} as a strong candidate, which has, to the best of our knowledge, only been used once in the domain of psychology and social science \parencite{VowelsLM2023attachment}. Super Learners are an ensemble approach, combining many machine learning approaches, each referred to as a candidate learner, within the same overarching algorithm. The Super Learner training process identifies an optimal weighted combination of predictions from each learner in order to arrive at a `consensus'. Importantly, the choice of learners should be diverse, including linear methods (such as linear regression), learners with limited flexibility (such as random forests with a low number of estimators), as well as highly flexible learners such as MultiLayer Perceptrons (MLPs; \cite{goodfellow}), and gradient boosting machines such as XGBoost \parencite{Chen2016b}. The weights for each of these learners are determined using a cross-validation scheme, which prevents a phenomenon known as overfitting, whereby the algorithms fail to approximate a function which generalises well to new samples. The motivation to include a diverse set of learners results in a very general prior on the function space, avoiding any bias towards a particular functional form.

The original proponents of the method explain that the Super Learner ensemble, with its specific cross-validated fitting process and diverse ensemble of candidate learners, yields favourable, almost parametric rates of sample efficiency.\footnote{Super Learners have $log(n)/n$ rate of convergence, compared with the parametric rate $1/\sqrt{n}$.} This means that whilst most individual machine learning algorithms require larger sample sizes for similar levels of convergence compared with parametric methods such as linear regression, Super Learners exhibit favourable performance in this regard. This is an important point, because whilst non-parametric estimators make fewer assumptions about the underlying process, this advantage is moot if they are otherwise inefficient and require disproportionately large sample sizes to achieve useful levels convergence.  Interested readers are directed towards recent work by \textcite{Rudolph2023, VowelsFreeLunch2022}, who examine the efficacy of non-parametric estimators in the context of causal inference. In particular, \textcite{Rudolph2023} come to the conclusion that certain types of estimator (including those which leverage the benefits of close-to-parametric rates of convergence like Super Learners), can indeed represent viable alternatives to linear estimators, even in the absence of large samples.

\begin{figure}[!ht]
\caption{SLEM / DAG Learner Block Diagram} 
\includegraphics[width=0.9\linewidth]{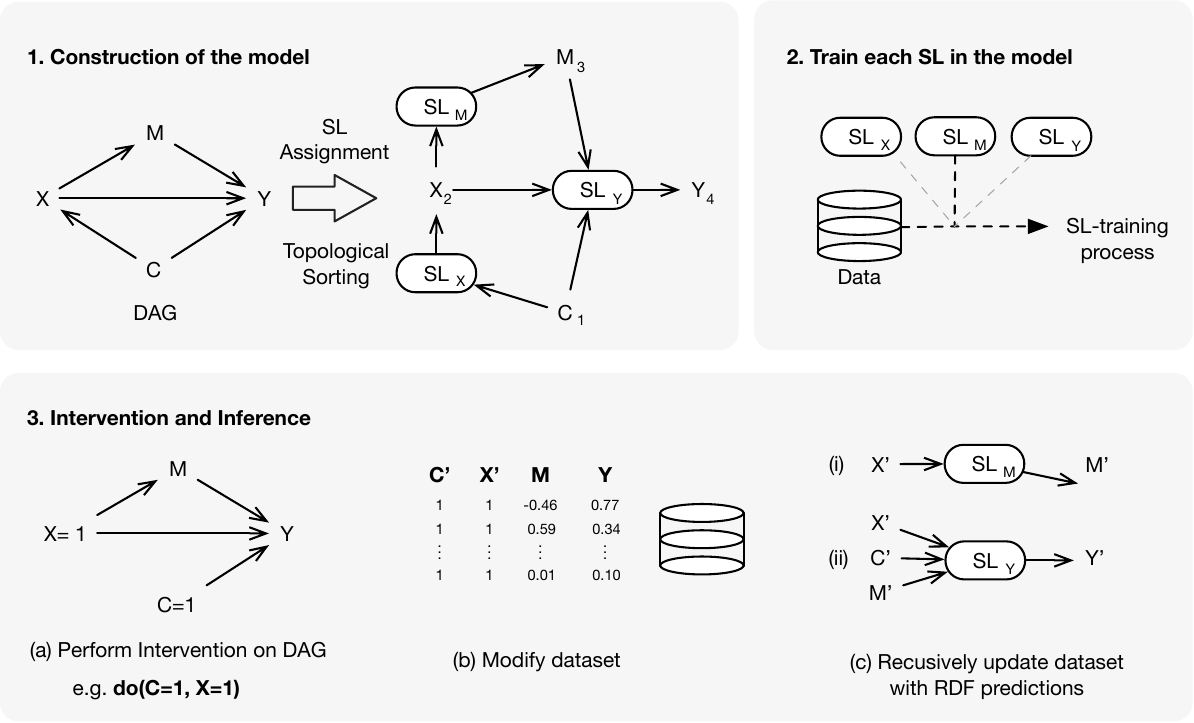}
\newline
\small{\textit{Note.} A block diagram of the SLEM process. (1) Firstly, a DAG is used to construct a model, where each variable with causal parents is assigned a Super Learner (SL). In the example, variables $X$, $M$, and $Y$ all have parents, and so each have a SL which will be used to map from the parents to their shared child/outcome. We have also labelled the causal ordering here ($C$ is the first variable in the causal ordering, and is therefore given a value 1, $X$ follows and has a value 2, and so on). (2) Secondly, each of the SLs is trained via a k-fold cross-validation process. (3) Thirdly, these SLs can be used to estimate the effects of multiple simultaneous interventions. In the example, interventions are performed on variables $C$ and $X$, setting them both to 1: $do(C=1, X=1)$. By consequence, their incoming arrows are removed, and the original dataset is updated with the corresponding values $C'$ and $X'$. Next, the structure is recursively iterated, using the modified dataset to obtain new predictor values and to store new outcome predictions. All non-intervention variables which are downstream of the intervention variables are updated, yielding an updated `intervention dataset' with all downstream variables modified according to the predictions from the associated SL.}
\label{fig:process}
\end{figure}

\section{Super Learner Equation Modeling (SLEM)}
\label{sec:slem}
Our framework is motivated by three requirements. Firstly, it must be able to perform causal inference. Secondly, it must be able to parse a user-specified theory in the form of a DAG. Thirdly, it must use machine learning to avoid assumptions about the functional form. We fulfil these requirements with SLEM, which combines DAGs with Super Learners (SLs) to facilitate estimation of all path coefficients in a DAG, as well as the estimation of arbitrary interventions (including multiple interventions simultaneously). The open-source code provides the SLEM DAG Learner object (accessible as a PyPI pip package \verb|slem-learn|, with repository here: \url{https://github.com/matthewvowels1/SLEM}). It is easy to install and to use, and the repository includes a tutorial style jupyter notebook \verb|example_usage.ipynb|. 

In this section, we begin by describing the overall process of the SLEM methodology (the steps are also depicted visually in Figure~\ref{fig:process}), before a discussion about its specific features and the DAG Learner.

\subsection{The Process}

It is worth noting that the use of Super Learners for causal inference is itself not new and has been long recommended in targeted learning literature \parencite{vanderLaan2014, vanderLaan2011, vanderLaanRubin2006, Coyle2020, Gruber2022}. Indeed, what we recommend is not fundamentally new \textit{per se}, but rather provides researchers with a generalisation of path modeling in an easy-to-use format. Indeed, most causal inference practice involves the specification of a single target `estimand', and the `identification' of this estimand (which is the process of expressing the causal quantity as something estimable from observed data) makes use of the DAG to establish which variables, other than the cause and effect variables, are necessary to arrive at a meaningful estimate. In contrast, approaches such as path modeling and SEM undertake the same process in an automatic fashion, and estimate all coefficients/effects associated with all paths in the model. Our goal with SLEM is to facilitate the same `DAG-wide' estimation process, making it possible to estimate all path coefficients, or, more generally, the result of any arbitrary hypothetical intervention, in one shot.

\subsubsection{Step 1: Construction of the Model}
\label{sec:step1}

Besides some very basic programming experience with Python, the user must also supply a dataframe containing the observed data and variable names, a networkX DAG \parencite{networkx}, and a Python dictionary containing the variable types for each variable (continuous, categorical, or binary). Such an example specification is shown in Application/Code Note 1. When the user supplies the DAG and variable types to the DAG Learner , a number of operations are undertaken automatically. The code required for the instantiation of the DAG Learner itself is also shown in Application/Code Note 2.

\begin{tcolorbox}
\textbf{Application/Code Note 1.}
Users are required to specify the following:
\begin{enumerate}
\item A pandas dataframe \parencite{pandas}:
\begin{lstlisting}[language=Python]
import pandas as pd
df = pd.read_csv('my_data.csv') \end{lstlisting}
    \item A networkX \parencite{networkx} directed graph object:\begin{lstlisting}[language=Python]
import networkx as nx
DAG = nx.DiGraph()
DAG.add_edges_from([('C', 'Y'), ('C', 'X'), ('X', 'Y')])  \end{lstlisting}
    \item A dictionary of variable types (`cont'inuous, `bin'ary or `cat'egorical):\begin{lstlisting}[language=Python]
var_types = {'C': 'cont', 'X': 'cont', 'Y': 'cont'}  \end{lstlisting}
\end{enumerate}
\end{tcolorbox}

\textbf{Topological Sorting:} Firstly, according to the user-specified DAG, a causal ordering is established. Each variable in the DAG is assigned a positive integer corresponding with its position in the causal ordering. For example, in the graph $X \rightarrow M \rightarrow Y$, variable $X$ is in position 1, $M$ in position 2, and $Y$ in position 3. It is perfectly possible for multiple variables to share a causal ordering. This `topological' / causal ordering is used to sort the variables and models, so that the model knows which variables have parents, and which variables should be updated in a response to particular intervention.

\textbf{Super Learner (SL) Assignment:} Secondly, each variable in the topologically sorted (\textit{i.e.}, sorted according to the causal ordering) DAG which has at least one parent is assigned as the outcome in a Super Learner prediction task. In other words, each variable with incoming causes is treated as an outcome in a prediction task, and an SL is used to make those predictions. It is possible that a single variable serves as a predictor for multiple SLs, if a variable is a parent for more than one variable. An example of this can be seen in Figure~\ref{fig:process}, where variable $C$ is a parent of both $X$ and $Y$, and is therefore used as a predictor in two SLs.
\begin{tcolorbox}

\textbf{Application/Code Note 2.}
Basic operation:
\begin{enumerate}
        \item The user can optionally specify a list of candidate learners for the Super Learner:
\begin{lstlisting}[language=Python]
learner_list = ['Elastic', 'LR', 'MLP', 'SV', 'AB', 'RF', 'BR', 'poly'] \end{lstlisting}
\item Instantiate a DAGLearner (note that the learner list argument can be omitted, and by default all learners will be used):\begin{lstlisting}[language=Python]
!pip install slem-learn
import slem
from slem import DAGLearner
daglearner = DAGLearner(dag=DAG,  var_types=var_types, k=6, learner_list=learner_list) \end{lstlisting}
    \item Fit the DAGLearner:\begin{lstlisting}[language=Python]
daglearner.fit(data=df) \end{lstlisting}
    \item Estimate all path coefficients:\begin{lstlisting}[language=Python]
ATEs = daglearner.get_0_1_ATE(data=df)  \end{lstlisting}
 \end{enumerate}
\end{tcolorbox}

\begin{algorithm}
\cprotect\caption{The DAG Learner \verb|infer| method for inferring the result of arbitrary interventions.}\label{alg:cap}
\begin{algorithmic}
\Require 
$\mathbf{X}$, $X \in \mathbf{X} \leftarrow$ Set of all graph variables,\\
$\mathbf{I} \subseteq \mathbf{X}$, $I \in \mathbf{I} \leftarrow$ Set of all intervention variables,\\
$\mathbf{V}$, $V \in \mathbf{V} \leftarrow$ Set of intervention variable values,\\
$\mathbf{O}$, $O \in \mathbf{O} \leftarrow$ Positive integers for causal ordering, \\
$\mathbf{Y} \subset \mathbf{X}$, $Y \in \mathbf{Y} \leftarrow$ Variables with causal parents, \\
$\mathbf{SL}$, $SL \in \mathbf{SL} \leftarrow$ Trained Super Learners,\\
$\mathcal{D} \leftarrow$ Empirical dataset,\\
$\mathbf{Q} = \cup_{I\in \mathbf{I}} \left(\mbox{dec}(I) \setminus \mathbf{I} \right) \leftarrow$ Set of all descendants of all intervention variables, not in the set of intervention variables,\\
$O : X \rightarrow O \;  s.t. \; O(X)$ returns the ordering for particular variable $X$, \\
$V: I \rightarrow V \; s.t. \; V(I)$ returns the value for intervention variable $I$,\\
$\mathcal{D}(X)$ are the empirical values for variable $X$,\\
$P : Y \rightarrow P \; s.t. \; P(Y)$ returns the causal parents of $Y$, \\
$SL : P(Y) \rightarrow \hat{Y} \; s.t. \; SL(P(Y))$ makes predictions $\hat{Y}$ from parents of $Y$ using empirical values $\mathcal{D}(P(Y))$ and using the $SL\in \mathbf{SL}$ which was trained to predict $Y$.
\For{$I \in \mathbf{I}$}
\State $\mathcal{D}'(I) := V(I) \leftarrow $ update dataset with corresponding intervention values.
\EndFor
\For{$Q_1, Q_2, ..., Q_K \in \mathbf{Q} \; s.t. \; O(Q_1) \leq O(Q_2) \leq ... \leq O(Q_K)$}
\State $\mathcal{D}'(Q) := SL(P(Q))$
\EndFor
\State\Return $\mathcal{D}' \leftarrow$ Dataset with modified values for interventional variables $V \in \mathbf{V}$ and descendants of these variables $\cup_{I\in \mathbf{I}} \left(\mbox{dec}(I) \setminus \mathbf{I} \right)$. 
\end{algorithmic}
\end{algorithm}

\subsubsection{Step 2: Super Learner Training}
\label{sec:step2}
Each of the SLs is trained using the user-supplied data and the associated predictors (causal parents) and outcome (causal child). Readers are directed to \textcite{Polley2007} and \textcite{vanderLaan2011} for a detailed description of the SL training process. In essence, for each SL, the data are split into $k$ folds. $k-1$ of these folds are used to train each learner separately in the SL, and the predictors in the final fold (the validation fold) are used to generate a set of predicted outcomes for each learner. Each learner is trained from scratch like this $k$ times, and each time a different fold is used to generate the predictions. 

Once one has $k$ sets of predictions from each learner, these predictions form what can be considered to be a new dataset, where each variable in this dataset constitutes the predictions from a particular learner. A `meta-learner' is trained on this dataset of predictions, where this meta-learner is usually a linear model with a weight for each learner, where these weights are constrained to be positive and to sum to one. These weights, derived on the validation fold predictions, determine to what extent each learner contributes to making a prediction. We expect learners which generalise well in terms of their validation predictions to have high weight, and those learners which perform poorly to have low weight. Finally, the all learners are retrained on the full dataset whilst retaining the individual learner-weightings derived according to the validation set.

\subsubsection{Step 3: Causal Estimation}
\label{sec:step3}
SLEM / DAG Learner has two operating modes for causal estimation. The first is called using the method \verb|daglearner.get_0_1_ATE(data)|. For each parent and each associated trained SL, the parent is first set to 0 and a set of predictions are generated. Next, the parent is set to 1 and a second set of predictions are generated. The two sets of predictions are subtracted from one another, and the average is taken. This provides an estimate for the Average Treatment Effect (ATE) of this particular causal parent for a particular associated causal child. More formally, for each variable $Y$ which has parents according to the DAG, the estimate (estimates indicated with $\hat{.}$ notation) for the ATE of $X$ on $Y$ can be expressed as:

\begin{equation}
    \hat{ATE}_{X\rightarrow Y} = \frac{1}{N}\sum_i^N \left( Q_Y(X=1, \mathbf{pa}(Y) \setminus X) - Q_Y(X=0, \mathbf{pa}(Y)\setminus X) \right).
\end{equation}

Here, $Q_Y$ is the SL for child variable $Y$, and $\mathbf{pa}(Y)  \setminus X$ denotes the set of parent variables of $Y$ not including $X$, where $X$ is the intervention variable being used for the estimation of the ATE of $X\rightarrow Y$ in the DAG. The DAG Learner code automates this procedure for all paths in the DAG, providing a set of estimated coefficients for each cause-effect pair.

The second operating mode for causal estimation is called using the method \verb|infer()|. An example usage is shown in Application/Code Note 3 and the corresponding algorithm is shown in Algorithm~\ref{alg:cap}. This method takes a user specified set of intervention variables and values (one value per intervention variable) and generates an `interventional dataset'.\footnote{Note that it is also trivial to integrate SLEM into a loop which iterates over multiple intervention values to estimate the effects of `soft' interventions.}  This dataset is generated by iteratively predicting the influence of the intervention on each variable in the DAG according to (a) whether or not they fall downstream of the intervention variable and (b) whether the variable is, itself, an intervention variable (in which case it will not be updated following an upstream intervention, because it has its incoming directed links removed). For instance, let's say one wishes to intervene on $X$ (the treatment) and $M$ (the mediator) in a partial mediation model involving $X$, $M$ and $Y$ (the outcome). The first step involves setting the values of $X$ and $M$ in the dataset to the desired intervention values. These values won't be updated, because we want to simulate an intervention, and therefore their incoming causal links in the DAG are effectively severed. The only two links which now exist are the links from $M$ and $X$ to $Y$. There is one SL modeling the incoming links to the outcome $Y$, and therefore the method only needs to generate one set of predictions in response to these two interventions. The trained SL for $Y$ is thus `fed' the dataset with the interventional values for $X$ and $M$, and it generates predictions $\hat{Y} | do(X=x, M=m)$. The resulting dataset provides the associated values for the estimated effect of the interventions.

\begin{tcolorbox}
\textbf{Application/Code Note 3.}
Custom Intervention/Inference:
\begin{enumerate}
        \item The user specifies their desired interventions. :
        \begin{lstlisting}[language=Python]
int_val_nodes1 = {'X': 1}
int_val_nodes0 = {'X': 0} \end{lstlisting}
\item Use the inference method to generate interventional datasets of predictions resulting from the interventions:\begin{lstlisting}[language=Python]
interventional_dataset1 = daglearner.infer(data=df, intervention_nodes_vals=int_val_nodes)
interventional_dataset0 = daglearner.infer(data=df, intervention_nodes_vals=int_val_nodes) \end{lstlisting}
    \item Compute the effect of a desired outcome variable:
\begin{lstlisting}[language=Python]
ATE = (interventional_dataset1['Y'] - interventional_dataset0['Y']).mean() \end{lstlisting}
 \end{enumerate}
 Note it is possible to perform multiple interventions simultaneously. For example: 
 \begin{lstlisting}[language=Python]
int_val_nodes = {'X': 1, 'C':0.5} \end{lstlisting}
This removes any influence from parents of $X$ and $C$, and sets their values to 1 and 0.5, respectively. DAGLearner will then take care of the inference according to the (interventionally modified) DAG.
\end{tcolorbox}

The output of the \verb|infer| method is at the participant level (\textit{i.e.}, the size of the interventional dataset is the same as the original), where each row technically represents an estimation of the Conditional Average Treatment Effect (CATE), where the conditioning is based on the other variables being used to make the predictions. This provides the researcher with the flexibility to explore (for example) group-specific effects in subsets of the data. If the researcher is simply interested in the average effect of the intervention, they can just take the difference between the two interventional datasets and take the average to compute the ATE.

\begin{figure}[!ht]
\caption{Using SLEM to explore moderation.} 
\includegraphics[width=0.7\linewidth]{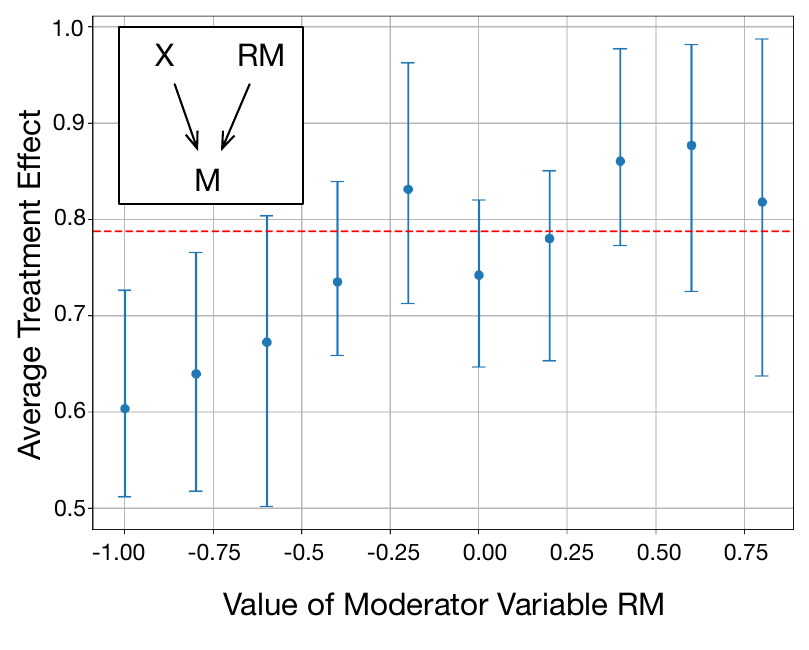}
\newline
\small{\textit{Note.} The advantage of the SLEM / DAG Learner \verb|infer| method is that one can iteratively pass multiple interventional values and use it to thereby estimate effects over different ranges of different variable values. This graph is the result of the combination of the \verb|infer| method and \verb|bootstrapper| function which we include in the software. For the case where the effect of $X$ on $M$ is mediated by a variable $R_M$, we estimate the difference in $M$ for $do(X=1, M=m_i)$ and $do(X=0, M=m_i)$ for $i \in [-1, 1]$. Associated code is provided in the \verb|example_usage.ipynb| tutorial notebook.}
\label{fig:bootstrapped_moderation}
\end{figure}

A demonstration of the strength of the \verb|infer| method is shown in Figure~\ref{fig:bootstrapped_moderation}, which depicts the estimated causal effects across a range of interventional values. In this plot, the red dashed line indicates the ATE of variable $X$ on mediator $M$, whilst the point estimations with confidence intervals indicate the estimated effect whilst intervening across a range of different values for a moderating variable $R_M$. In this way, one can explore how causal effects change non-linearly for different combinations of interventions.

\subsection{SLEM Features}
\label{sec:features}

\subsubsection{Super Learner Options} 

The SLEM DAG Learner object comes with the following included candidate learners for either regression or binary or multiclass classification: linear and logistic regressors, ElasticNet \parencite{Zou2005}, MultiLayer Perceptrons \parencite{goodfellow}, Support Vector Machines \parencite{Platt1999, murphy2}, AdaBoost \parencite{AdaBoost}, Random Forests \parencite{breiman2001}, Bayesian Ridge Regression or Naive Bayes Classification \parencite{ZhangBayes, Tipping2001}, and polynomial features regression. The implementations in scikit-learn library \parencite{sklearn} are used for each learner. Currently, the default hyperparameter settings including the scikit-learn library are used (at least for some algorithms, these settings have been shown to work well; \cite{Probst2019}), although our framework can be adapted simply to include, for example, multiple random forests with a range of different settings.

If the user simply wishes to implement a linear/logistic model, they can set the argument \verb|baseline=True| when instantiating a DAG Learner object.

\subsection{Bootstrapping}

We provide a bootstrapping \parencite{Efron1981, efron} function to provide researchers with direct access to confidence intervals and statistical inference. Example usage is shown in Application/Code Note 4. The bootstrapping function accepts three key arguments: The number of bootstrapped sub-samples to perform, the sample size for each subsample, and the dataset to bootstrap over. Otherwise, it requires the same arguments as when instantiating a DAG Learner, because within each bootstrap it itself instantiates a new learner. It can also accept either one desired dictionary of intervention nodes and values, or two if a contrast is desired.

\begin{tcolorbox}
\textbf{Application/Code Note 4.}
Bootstrapping:
\begin{enumerate}
        \item For estimation of all path coefficients:
        \begin{lstlisting}[language=Python]
bs_results_ATE = slem.bootstrapper(num_bootstraps=10, subsample_size=bootstrap_subsample_size, k=k, data=df, dag=DAG, var_types=var_types) \end{lstlisting}
\item For estimation of a specific, user-specified intervention: 
\begin{lstlisting}[language=Python] 
int_nodes_val = int_val_nodes = {'X': 0}
bs_results_int = slem.bootstrapper(num_bootstraps=10, subsample_size=bootstrap_subsample_size, k=k, int_nodes_val=int_nodes_val, data=df, dag=DAG, var_types=var_types) \end{lstlisting}
    \item For estimation of a specific cause-effect contrast: \begin{lstlisting}[language=Python]
int_val_nodes = {'X': 0}
int_nodes_valb = int_val_nodes = {'X': 1}
bs_results_contrast = slem.bootstrapper(num_bootstraps=10, subsample_size=bootstrap_subsample_size, k=k, int_nodes_val=int_nodes_val, int_nodes_valb=int_nodes_valb, data=df, dag=DAG, var_types=var_types \end{lstlisting}
 \end{enumerate}
\end{tcolorbox}

\subsection{Fit/Prediction Metrics}

For each outcome identified by SLEM, a set of out-of-sample performance metrics are provided depending on the variable type. For continuous variables, SLEM provides mean absolute error, mean squared error, median absolute error, R-squared and variance explained. However, note that R-squared and variance explained are not reliable in non-linear models, owing to the possibility of (amongst other things) non-homogeneity of variance. As such, we advise that users interpret these latter two fit statistics with caution. For binary or categorical variables, SLEM provides an overall accuracy score, an overall balanced accuracy score (which helps to account for any class or category support imbalance), precision, recall, and F1 score.

\section{Simulations}
\label{sec:sim}
The purpose of these simulations is (a) to verify that our proposal SLEM provides consistent estimation of the true causal effect size, (b) to demonstrate the importance of machine learning in the context of causal inference and to verify that SLEM does indeed perform well regardless of whether the underlying Data Generating Process (DGP) is linear or non-linear, and (c) to explore the sample efficiency of SLEM compared with SEM in practice. Additional simulation results are provided in the Appendix/Supplementary. Presently, we provide the results for three important simulations depicted in Figures~\ref{fig:nonlin}, \ref{fig:simple_linear}, and \ref{fig:simple_nonlinear}. 

\subsection{Simulation Construction}
For all sets of simulations, the SLEM DAG Learner uses all 8 currently included candidate learners. For the first simulations, the results in Figure~\ref{fig:nonlin} were obtained using a sample size of 10,000, by generating data with varying levels of non-linearity, and repeating the simulation 20 times to obtain average estimates of the importance of $X_1$ on $Y$ (which in actuality is zero). The DGP for this process is given below:

\begin{equation}
    \begin{split}
        U_Y \sim \mathcal{N}(0,1), \; \; U_{X_1} \sim \mathcal{N}(0,1), \; \;  U_{X_2} \sim U(-10, 10)\\
        X_1 := U_{X_1}, \; \; X_2 := 0.5X_1 + U_{X_2}\\
        Y = \lambda_1 X_2 + \lambda_2 X_2^2 + \lambda_3 X_2^3 + U_Y
    \end{split}
\end{equation}

 Here, $\sim$ indicates samples are randomly drawn, $\mathcal{N(0,1)}$ denotes a standard normal distribution, $U(-10, 10)$ denotes a random uniform distribution ranging from negative to positive 10, and $\lambda$ represent a set of coefficients on the third degree polynomial features of $X_2$. The varying degree of non-linearity relating $X_2$ to $Y$ is achieved by varying the $\lambda$ coefficient value such that $\lambda_2$ and $\lambda_3$ begin around zero and increase linearly. The data are used to fit a SLEM DAG Learner as well as a multiple linear regression (using the linear regressor included in scikit-learn), and the figures depict the resulting distribution of Mean Absolute Errors (MAEs) for the estimation of the effect size of $X_1$ on $Y$ for each method across each sample size. The MAE tells us how close the estimated effect sizes are to the truth effect size, and their distribution also tells us how much this error varies around the average predicted value.

Secondly, the two sets of simulation results depicted in the Figures~\ref{fig:simple_linear} and \ref{fig:simple_nonlinear} are constructed by first generating 70 datasets according to the associated DAG/DGP and an associated set of SCM equations. This is then repeated for six different sample sizes: 50, 100, 250, 500, 1000, 5000, resulting in 420 datasets. Finally, results are obtained for the MAE of estimation for the effect of $X$ on $Y$ using the SLEM DAG Learner and compared against the effect size estimations obtained using a SEM \verb|lavaan| R library \parencite{Rosseel2012} specified (structurally) correctly according to the DAG. The DGPs for the linear and non-linear DGPs are:

\begin{equation}
    \begin{split}
        U_X \sim \mathcal{N}(0,1),\; \; \;  U_Z \sim \mathcal{N}(0,1), \; \; \;  U_Y \sim \mathcal{N}(0,1),\\
        Z := U_Z, \; \; \;   X \sim Bi(p=\sigma(0.7Z + U_X)),\\
        Y := -0.7X + 0.8Z + U_Y,\\
    \end{split}
\end{equation}

and 

\begin{equation}
    \begin{split}
        U_{Z_1} \sim \mathcal{N}(0,1),\; \; \; U_{Z_2} \sim \mathcal{N}(0,1), \; \; \; 
        U_X \sim \mathcal{N}(0,1), \; \; \;  U_Y \sim \mathcal{N}(0,1),\\
        Z_1 := U_{Z_1},\; \; \;  Z_2 := U_{Z_2},\\
        X \sim Bi(p=\sigma(0.7Z_1 + 0.4Z_1Z_2 + 0.4Z_2 + U_X)),\\
        Y := 0.3 + 1.5X + 0.8Z_1 + 0.3Z_1Z_2 + 0.5Z_2^2 + U_Y,\\
    \end{split}
\end{equation}

respectively. Here, $Bi(p)$ indicates a binomial distribution, and $\sigma$ indicates the sigmoid function.

\subsection{Results: Consistent and Unbiased Estimation for Linear and Non-Linear DGPs}
\begin{figure}[!ht]
\caption{Simple linear SEM vs. the SLEM DAG Learner comparison simulation.} 
\includegraphics[width=0.8\linewidth]{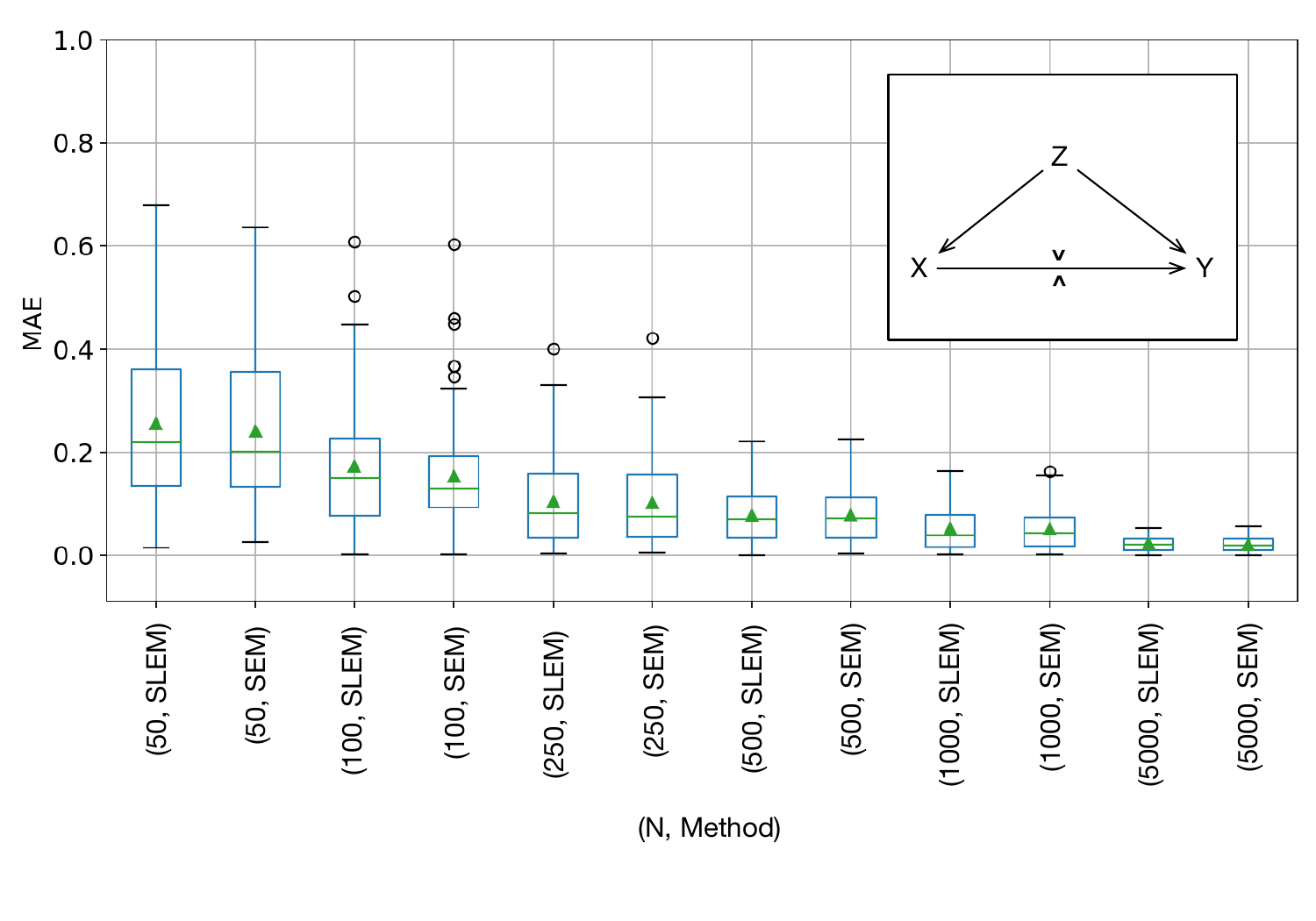}
\newline
\small{\textit{Note.} Mean Absolute Error (MAE) for the estimation of the true effect of $X$ on $Y$ by SLEM / DAG Learner and SEM, for data generated \textit{linearly} according to the DAG (inset for convenience) across a range of sample sizes. Green arrows represent the average MAE, whilst the horizontal lines depict the median.}
\label{fig:simple_linear}
\end{figure}

An estimator is consistent if its estimates converge as sample size increases, and unbiased if its average is centred around the true value. We actually already know the estimates for both SEM and SLEM are unbiased in the linear case (which applies to the results in Figure~\ref{fig:simple_linear}), due to our integration of the causal DAG framework with $do$-calculus. In other words, we are able to express the true causal effect as a function of the observational data because we have access to the true structure used to generate the data. Furthermore, SEM is specifically unbiased in terms of its functional misspecification, because it is linear and so is the DGP. 

Both the unbiasedness and the consistency of SLEM and SEM are illustrated in Figure~\ref{fig:simple_linear}, by virtue of the decrease in the average error (indicated by the green arrow) which approaches zero as sample size increases. It is worth acknowledging that SEM exhibits slightly lower average MAEs than the SLEM DAG Learner, although (a)  this is unsurprising given it is exactly the correct model (both structurally and functionally), and (b) this difference decreases as sample size increases. In practice, of course, both methods may end up being structurally misspecified (because our DAG represents our current state-of-knowledge regarding the development of a simplified model representing reality), whilst the SEM is likely to be both structurally \textit{and} functionally misspecified owing to its assumption of linearity.

\begin{figure}[!ht]
\caption{Simple non-linear SEM vs. the SLEM DAG Learner comparison simulation.} 
\includegraphics[width=0.8\linewidth]{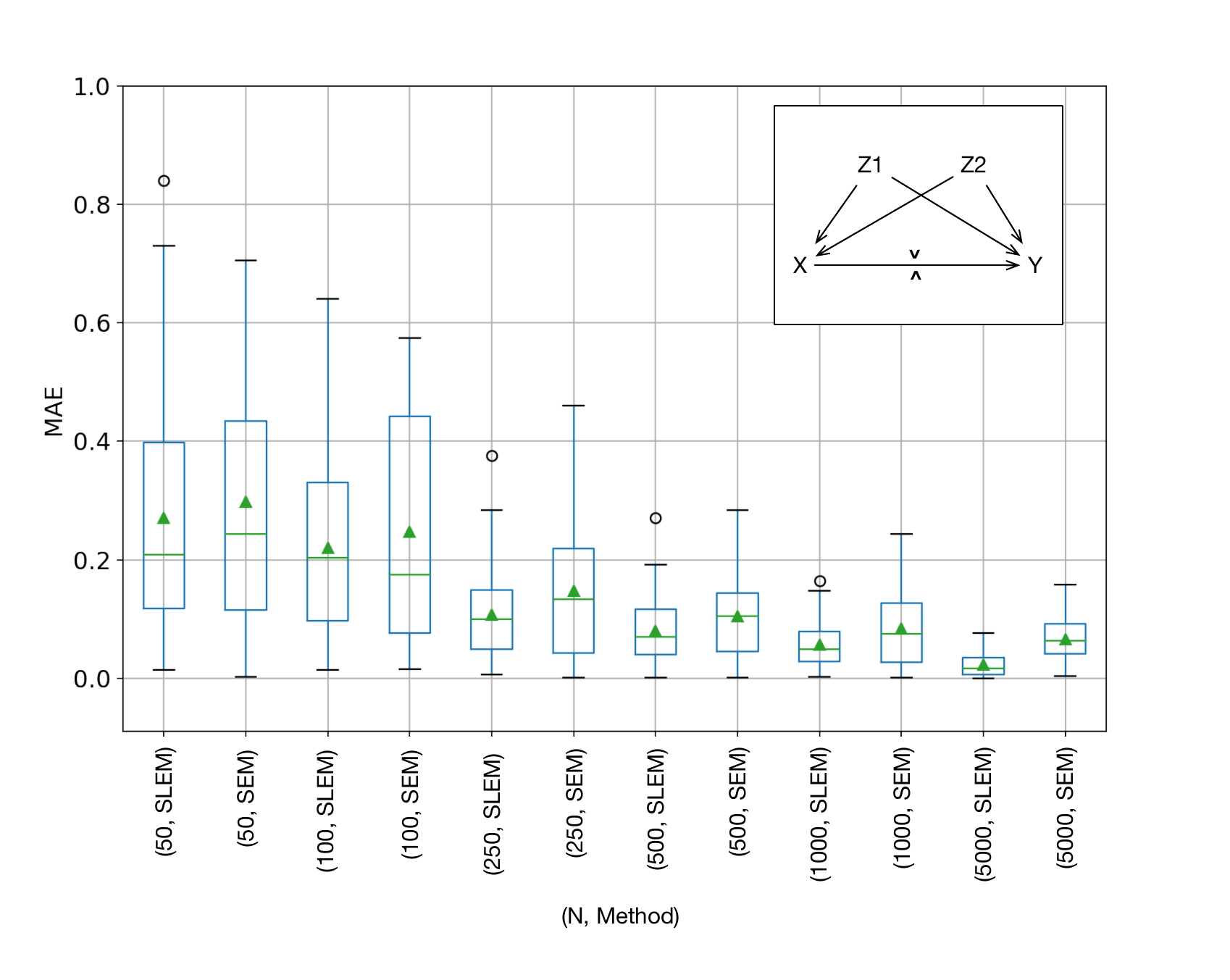}
\newline
\small{\textit{Note.} Mean Absolute Error (MAE) for the estimation of the true effect of $X$ on $Y$ by SLEM / DAG Learner and SEM, for data generated \textit{non-linearly} according to the DAG (inset for convenience) across a range of sample sizes. Green arrows represent the average MAE, whilst the horizontal lines depict the median.}
\label{fig:simple_nonlinear}
\end{figure}

In Figure~\ref{fig:simple_nonlinear}, which presents the results for the non-linear case, we see exactly the consequence of achieving structural specification but incorrect functional specification. Whilst the MAE of SLEM tends to zero (and is therefore both unbiased and consistent in this non-linear case), the MAE of SEM does not, because it is now functionally misspecified. Therein lies the advantage of SLEM insofar as it has the same risk of structural misspecification as SEM, but a far reduced risk of functional misspecification.

\begin{figure}[!ht]
\caption{Non-Linear Simulation} 
\includegraphics[width=0.7\linewidth]{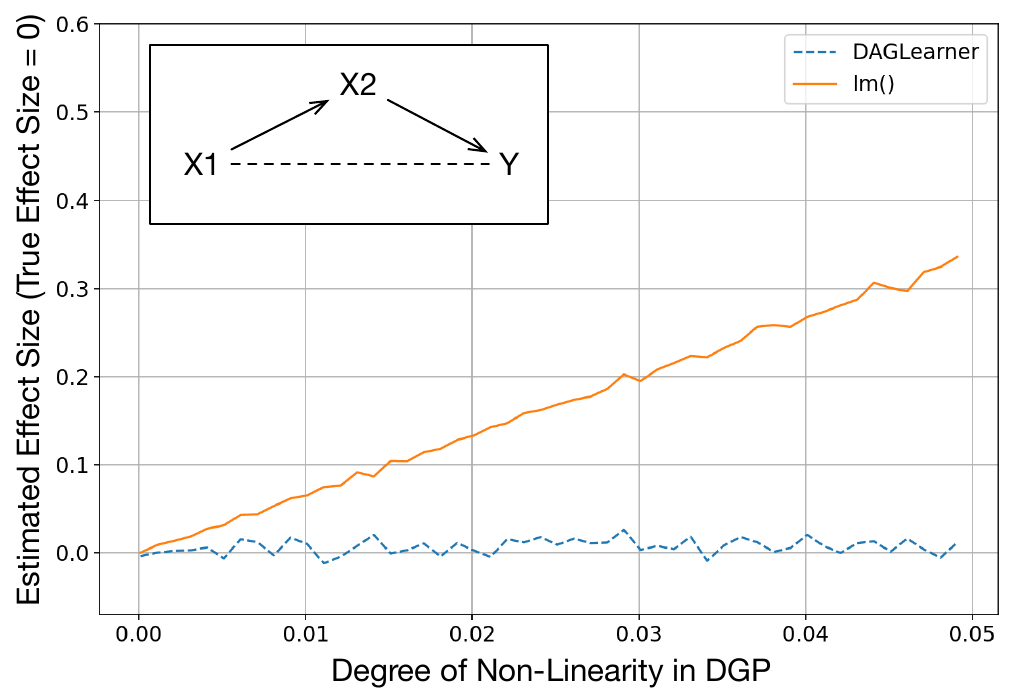}
\newline
\small{\textit{Note.} These simulations are for a very simple DAG $X_1 \rightarrow X_2 \rightarrow Y$, and show that the SLEM DAG Learner can adapt to increasing degrees of polynomial non-linearity successfully, whereas linear models cannot. More specifically, $X_2$ fully mediates the effect of $X_1$ on $Y$, and therefore, conditional on $X_2$, there is no (direct) effect of $X_1$ on $Y$. This zero effect should be recoverable using linear estimation methods (either SEM or linear regression), but we introduce polynomial non-linearity in the relationship between $X_2$ and $Y$ which varies in intensity according to higher values of the coefficients on the polynomial features of $X_2$ (x-axis). As a result, linear regression wrongly assigns importance to the otherwise conditionally independent variable $X_1$. In contrast, SLEM correctly accounts for the increase in non-linearity, and consistently and correctly estimates the effect of $X_1$ on $Y$ given $X_2$ to be zero.}
\label{fig:nonlin}
\end{figure}

More generally, the importance of machine learning in the context of causal inference is demonstrated in Figure~\ref{fig:nonlin}, which depicts the estimated importance / causal effect size of a variable which is conditionally independent of the outcome. When the amount of non-linearity in the DGP is low (on the left-hand side of the plot) both the SLEM DAG Learner and a linear regression correctly estimate the the effect of $X_1$ on $Y$ to be zero. However, as the degree of non-linearity increases, the linear regression begins to assign importance to $X_1$, as if there existed a direct effect of $X_1$ on $Y$, whilst the SLEM DAG Learner, on the other hand, correctly continues to estimate the effect size to be zero.

\subsection{Results: Empirical Sample Efficiency}
Whilst a more theoretical evaluation of sample efficiency is provided in the original Super Learner proposal \parencite{Polley2007}, the practical performance will vary based on the data and the choice of candidate learners and their diversity. To this extent, the sample efficiency of SLEM and SEM can be empirically evaluated by reviewing the intervals of MAE in the simulation results shown in Figures~\ref{fig:simple_linear} and \ref{fig:simple_nonlinear}. Interestingly, the variation produced by SEM is not markedly less than SLEM, and in some cases appears to be equivalent or even marginally greater (see Figure~\ref{fig:simple_linear} when N=250 and N=500, for example). Of course, in the non-linear case, we expect the variation to appear greater for SEM because it is functionally misspecified. Despite this, the variation in the error itself is nonetheless notably large (see Figure~\ref{fig:simple_nonlinear} when N=250, for example). These results indicate that the use of SLs does indeed yield competitive sample-efficiency compared with parametric approaches, and this positive result may be due to the inclusion of the diverse range of candidate learners, in particular the inclusion of linear/logistic regressors.

\section{Benchmark Dataset Evaluation}
We also evaluate SLEM using the IHDP dataset \parencite{Hill2011, Dorie2016}\footnote{Available from \url{https://www.fredjo.com/}}, which has served as a common benchmark for causal inference tasks. Note that the use of such benchmark datasets for causal inference has been recently criticized on the basis that their use tends to result in the development of methods which overfit on these datasets, and that it is instead better to design bespoke datasets which are intended to specifically test the unique characteristics of a method \parencite{VowelsFreeLunch2022, Curth2021b}. Furthermore, we note that Super Learners, as an estimation method, have already been extensively evaluated elsewhere and as such we provide this comparison primarily as a validation check that the package is functioning as expected, and for completeness.

We use the version of IHDP corresponding with setting A of the NPCI package \cite{Dorie2016} (see \cite{Shi2019, Shalit2017}, and \cite{Yao2018}) and comprises 608 untreated and 139 treated samples (747 in total).\footnote{This variant actually corresponds with what is referred to as variant B in \parencite{Hill2011}.} There are 25 covariates, 19 of which are discrete/binary, and the rest are continuous. The outcome generating process is designed such that under treatment, the potential outcome is exponential, whereas under no treatment the outcome is a linear function of the covariates \parencite{Curth2021b}.

In order to evaluate and compare the performance of SLEM against other methods, we employ two commonly used metrics - the error on the estimation of ATE `eATE', and the error for the Precision in Estimating Heterogeneous Effects `ePEHE'. These metrics are defined as follows:

\begin{equation}
    e_{ATE} = | \hat{ATE}_{T\rightarrow Y} -  ATE_{T\rightarrow Y}  |, 
\end{equation}

\begin{equation}
    e_{PEHE} = \sqrt{\frac{1}{N}\sum_{i=1}^{N}\left( \hat{ATE}_{T\rightarrow Y, i } - ATE_{T\rightarrow Y, i }\right)^2} ,
\end{equation}

where the former is essentially the absolute error on the estimation of the average treatment effect, and the latter is a form of root-mean-squared-error across the conditional treatment effects $\hat{ATE}_{T\rightarrow Y, i }$ estimated for each individual $i$ in the dataset.

SLEM is compared against a range of recent (predominantly neural network based) methods for causal inference, and the results are shown in Table~\ref{tab:IHDPresults}. The results show competitive performance with state-of-the-art methods for causal inference, particularly given that many of the alternative methods involve hyperparameter tuning, whereas SLEM is run only once with its default configuration.

\begin{table*}[ht!]
\centering
\caption{Means and standard errors for evaluation on IHDP \cite{Hill2011}. Results from: \cite{Louizos2017b, Shalit2017,Zhang2020, Yoon2018}. `oos' is out-of-sample and `ws' is within sample. Lower is better for all metrics.}
\begin{tabular}{lllll} 
\\
\textbf{Method} & $\sqrt{\epsilon_{PEHE}}$ ws & $\sqrt{\epsilon_{PEHE}}$ oos & $\epsilon_{ATE}$ ws &$\epsilon_{ATE}$ oos \\ \hline
BART \cite{ChipmanBART}       & 2.1$\pm$.10  & 2.3$\pm$.10  & .23$\pm$.01  & .34$\pm$.02  \\
CEVAE \cite{Louizos2017b}       & 2.7$\pm$.10  & 2.6$\pm$.10  & .34$\pm$.01  & .46$\pm$.02  \\
TARNet \cite{Shalit2017}       & .88$\pm$.00  & .95$\pm$.00  & .26$\pm$.01  & .28$\pm$.01  \\
CFR-MMD \cite{Shalit2017}      &  .73$\pm$.00 & .78$\pm$.00  & .30$\pm$.01  & .31$\pm$.01  \\
CFR-Wass \cite{Shalit2017}     &  .71$\pm$.00 & .76$\pm$.00  & .25$\pm$.01  & .27$\pm$.01 \\
TEDVAE \cite{Zhang2020} & .62$\pm$.11 &  .63$\pm$.12 &- & .20$\pm$.05  \\
IntactVAE \cite{Wu2021} & .97$\pm$.04 & 1.0$\pm$.05  & .17$\pm$.01&  .21$\pm$.01 \\
GANITE \cite{Yoon2018} & 1.9$\pm$.40 & 2.4$\pm$.40 & .43$\pm$.05 & .49$\pm$.05 \\
Dragonnet w/ t-reg \cite{Shi2019} & - & - & .14$\pm$.01 & .20$\pm$.01 \\
TVAE \cite{Vowels2020b} & .52$\pm$.02 & .54$\pm$.02 & .15$\pm$.01 & .16$\pm$.01 \\
\textbf{SLEM}  &  .87$\pm$.66 & .91$\pm$.78 & .11$\pm$.10 & .17$\pm$.14 \\
\hline
\end{tabular}
\label{tab:IHDPresults}
\end{table*}

\section{Assumptions, Limitations, and Alternatives}
\label{sec:assumps}
The four principal limitations of SLEM, in particular with respect to the linear alternative SEM, are (1) the computation time required to fit the SLs, (2) the lower sample-efficiency / slower convergence rates, (3) the need to undertake some kind of non- or semi-parametric technique for performing statistical inference, and (4) the fact it does not handle latent variables. We discuss these in turn, as well as commenting on the absence of overall model-fit statistics, and a presentation of alternatives to SLEM.

\subsection{Computation Time} 
In respect to the first limitation, it takes 8.6 seconds to fit a SL to a dataset with 600 samples and 14 variables on a PC with a i9-9900K CPU. Thus, if one wishes to undertake bootstrapping (which requires fresh instantiation of the SLEM DAG Learner for each bootstrap), one would need around 15 minutes for 100 bootstraps. Of course, this scales according to the number of variables, the sample size, and the number of causal children in the DAG. 

\subsection{Sample Efficiency} 
With respect to the lower sample-efficiency and slower convergence rates, we have already noted that SLs have favourable rates of convergence compared to most other machine learning methods, but indeed, one expects an increased degree of estimation variation compared with linear methods. This notwithstanding, in practice, our simulations indicate sample efficiency not markedly dissimilar, and in some cases equivalent or better, compared to the sample efficiency of SEM.

\subsection{Statistical Inference} 
In the associated code we provide a bootstrapping function for deriving confidence intervals for the purposes of undertaking statistical inference with the SLEM approach (such as null-hypothesis significance testing). Unfortunately, as mentioned above, bootstrapping adds to the computational burden associated with the SLEM methodology. Indeed, it is worth remembering that when methods such as SEM or linear regression are used, statistical inference often follows rather conveniently as a consequence of the strong parametric or functional assumptions associated with these methods. Besides the bootstrapping approach, researchers can also consider alternatives for undertaking statistical inference when implementing non-parametric models, such as semi-parametric methods \parencite{VowelsFreeLunch2022, Robins2008, Ichimura2015, Hines2021}.

It is possible that the sample efficiency would decrease if users were to create a long chain of mediators, intervene on the first variable in this chain, and estimate its effect on the last variable in this chain. In this case, the \verb|infer| method would (according to its normal operation) iteratively predict the effect of the intervention on all subsequent mediating variables, and the associated error could lead to a compounding error which propagates down the length of the chain. In our view, this would not represent `good practice', and users should simplify their graph according to their particular research questions (indeed, if the chain of events concerning the mediators is not of interest, they can be removed from the DAG; readers can consult \cite{Vowels2023_prespec} for more information).

\subsection{Latent Variables}
The traditional SEM framework is well known for undertaking not only the estimation of observed or structural path coefficients in what is often referred to as the \textit{structural}, but also for estimation latent variables as part of a \textit{measurement} model \parencite{Kline2005}. In contrast, SLEM requires all the variables in the graph to be observed, and cannot perform latent inference. As such, its primary advantage and application area lies in its potential as a path/structural model. Researchers are required to undertake latent inference before using SLEM in order to utilize latent variables as observed variables in the graph, and are free to use a method of their choosing to achieve this (see also the discussion below on alternatives).

\subsection{Fit Statistics} 
In traditional Structural Equation Modeling, it is typical to provide a range of fit statistics associated with a given model. These include the Root Mean Squared Error of Approximation, Chi-square statistic, Comparative Fit Index, the Tucker-Lewis Index, or the Akaike's and Bayesian Information Criteria. We make a conscious decision not to provide such overall measures of model fit, for a number of key reasons. Firstly, fit statistics should not be used to guide modeling decisions that concern causal theory \cite{VowelsPipeline, Pearl2009}. Indeed, it is well known that `fitness to data is an insufficient criterion for validation causal theories' \cite{Pearl2009}, and it is quite possible to have a correctly specified model that has a worse set of fit statistics than a completely incorrectly specified model. Secondly, the majority of commonly used fit statistics are parametric, and therefore are not directly transportable to a completely non-parametric, non-linear approach such as SLEM. Even the general statistics such as AIC and BIC (which are expressed in terms of log-likelihood scores) are non-trivial to adapt to models with a mixture of continuous, binary, and categorical data types, and require their own set of generally sample-inefficient, multivariate, density estimation methods. Finally, and as described in the section above on SLEM features, SLEM already provides a range of predictive performance statistics for each individual outcome modeled by a Super Learner (including mean square error and F1 score). In the event that a researcher really wants to use fit statistics to, for instance, select between two models, we advise the user interprets the fit statistics associated with the specific effect size(s) of interest, and prioritize the model with a higher predictive capacity for modeling the proximal relationships relevant to the target effect(s). Such targeted interpretation would not be possible if one only had access to overall measures of model fit, which do not allow one to assess the specific success of a set of Super Learner submodels pertinent for the estimation of a particular causal effect. Note, even then, that such use of fit statistics for such causal model selection is not advised, and we would instead recommend researchers integrate domain knowledge and expertise with causal discovery techniques for a more principled approach. For an overview of causal discovery methods (including those for non-parametric data), readers are directed to the review by \cite{Vowels2021DAGs} and the theory development pipeline presented in \cite{VowelsPipeline}.

\subsection{Alternatives} 
Besides the class of SEM approaches which require users to specify the functional form relating variables \textit{a priori} \parencite{Marsh2006}, the space of approaches for non-linear and non-parametric SEM is quite busy. For instance, there are methods which fit non-linear relationships from data, but make assumptions about the parametric form of the variables. This includes the Latent Moderated Structural Equations (LMS) approach which assumes the variables themselves are normally distributed \parencite{Klein2000}. Additionally, there exists a class of models known as  semiparametric structural equation mixture models (SEMMs), which combine mixtures of linear SEMs in order to approximate non-linear relationships when the latent predictors are not normally distributed (see, for example, \cite{Pek2011, Kelava2014}). The two key advantages of this latter approach is that they can be used to estimate latent variables, and that it is not necessary to make assumptions about the parametric or functional form. Unfortunately, they do not provide effect size estimation (although could probably be extended to do so) \parencite{Umbach2017}. The nonlinear structural equation mixture modeling (NSEMM) approach \cite{Kelava2014} provides such effect size estimates, and makes no assumptsions about the normality of the latent variables, but is limited in terms of its estimation methods. Perhaps the most flexible alternative to our proposal is provided in the \verb|nlsem| package \parencite{Umbach2017}, which combines SEMM, LMS and NSEMM, thereby providing parameter estimates, handling latent variables, and making no functional or parametric assumptions. However, whilst these methods profit from their capacity to model latent variables, they are limited according to specific estimation methods (\textit{e.g.}, the use of Gaussian mixture models, or in the case of NSEMM, the assumption of normally distributed residuals). Finally, the framework presented by \parencite{vanKesteren2022} re-conceptualizes SEM in the context of computational graphs in order to open the door to flexible developments in the area. Furthermore, alternative state-of-the-art machine learning approaches to causal inference such as semi-parametric targeted learning approaches \parencite{vanderLaan2011, VowelsLM2023attachment}, neural network based approaches \parencite{Vowels2020b, Zhang2020, Louizos2017b}, or theory-constrained decision trees and random forests \parencite{Brandmaier2016,Brandmaier2013} are, at least in their distributed forms, either challenging to understand and utilize, limited to specific estimators (the exception is targeted learning based approached which incorporate Super Learners), or limited to the estimation of specific effects and do not provide general estimation of all include paths and all possible interventions.

As such, to the best of our knowledge, there are no alternative methods which (a) avoid unnecessary assumptions about the functional form, (b) provide close to parametric rates of convergence / sample efficiency, (c) provide estimation of all causal effects in a DAG, (d) facilitate the estimation of causal effects from arbitrary interventions, and (e) are easy to use. Even though \verb|nlsem| \parencite{Umbach2017} represents a flexible alternative which (in contrast to SLEM) handles latent variables and which is compiled into an easy-to-use R package, it does not (at least by default) fulfil (d), and thereby does not enable users to specify arbitrary interventions. Indeed, whilst SLEM is limited to the \textit{structural} elements of a structural causal model (and cannot handle latent variables), it benefits from near-parametric rates of convergence and practically no assumptions concerning the functional or parametric form. SLEM reflects a particularly flexible and `causal philosophy' in this regard. 

We would also argue that the ease-of-use element (e) should not be understated, and packages or methods which are not easy for applied researchers present their own barriers to adoption by applied researchers. Indeed, it is our view that the computational graph approach of \textcite{vanKesteren2022} is quite technical and mostly of benefit to statisticians and specialists in statistics, machine learning, and SEM. More generally, and in spite of the aparent abundance of flexible approaches to SEM, most statistical models, particularly in psychology and social science, tend to involve linear and parametric assumptions \parencite{Blanca2018}. At the very least, applied practitioners seem to have been slow to integrate these approaches into their toolboxes, and this perhaps speaks to the need for simple packages which are easy for non-specialists to integrate into their research pipelines.

\section{Conclusion}
\label{sec:Conclusion}
In this paper, we present an integrated framework and easy-to-use code which enables researchers to estimate the causal result of any desired intervention (including multiple possible interventions simultaneously) on a user-specified causal Directed Acyclic Graph. The method integrates Super Learners, a state-of-the-art machine learning ensemble method with favourable convergence rate, thereby avoiding any unnecessary assumptions about the underlying functions relating the variables included in the DAG. Similar results were obtained a discussed in recent work by \textcite{Rudolph2023}, examining the relevance of non-parametric techniques for estimation. Our simulation results confirm that SLEM has a performance which is both highly competitive compared with conventional Structural Equation / Path Modeling on data generated according to linear Data Generating Processes, and clearly outperforms such methods if there exist non-linear relationships.

\printbibliography

\newpage
\appendix

\section{Supplementary Material}
\subsection{Additional Simulations}
\begin{figure}[!ht]
\caption{Complex linear SEM vs. the SLEM DAG Learner comparison simulation.} 
\includegraphics[width=0.8\linewidth]{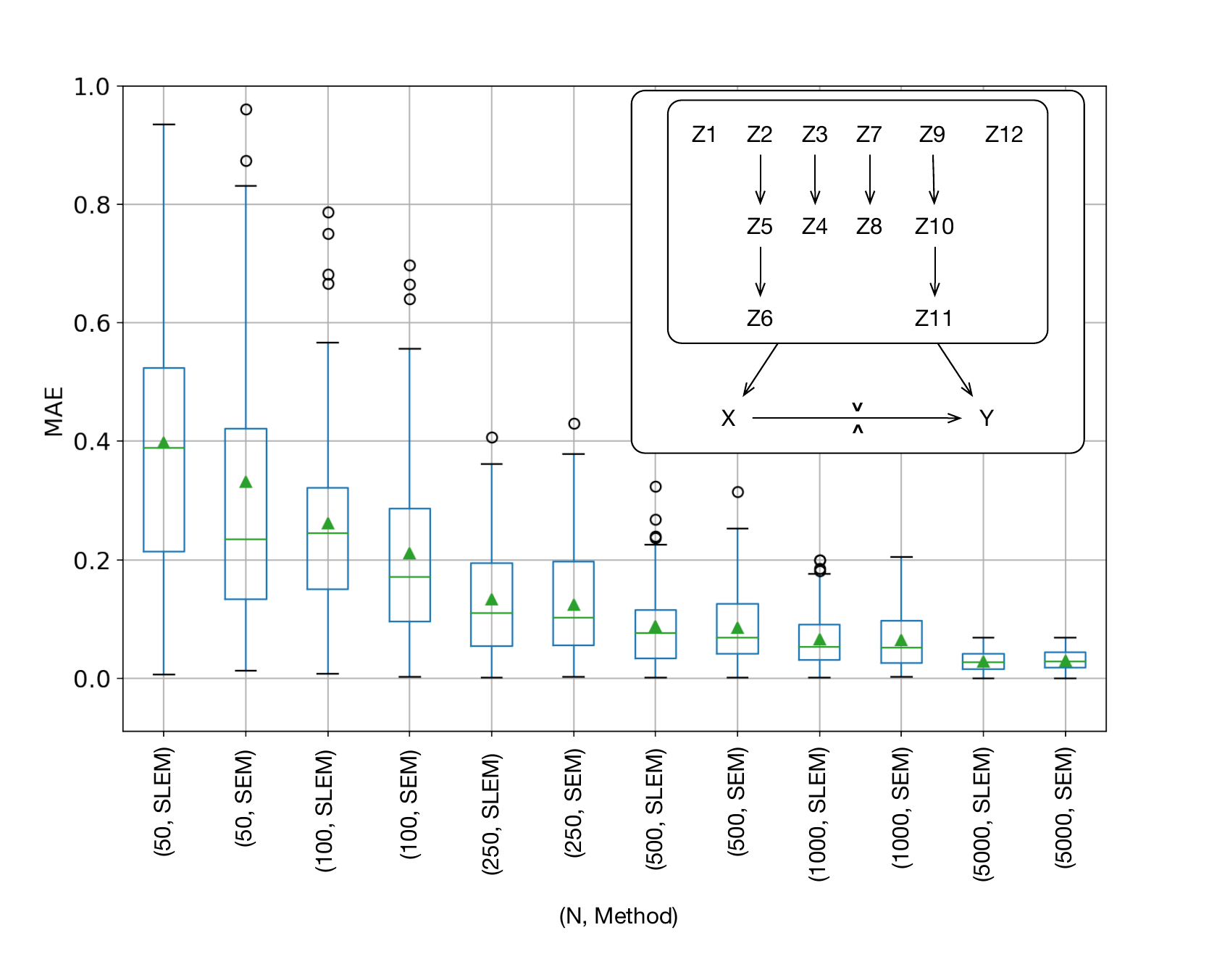}
\newline
\small{\textit{Note.} Mean Absolute Error (MAE) for the estimation of the true effect of $X$ on $Y$. Green arrows represent the average MAE, whilst the horizontal lines depict the median.}
\label{fig:DGP_B}
\end{figure}

\begin{figure}[!ht]
\caption{Simple partial mediation SEM vs. the SLEM DAG Learner comparison simulation.} 
\includegraphics[width=0.8\linewidth]{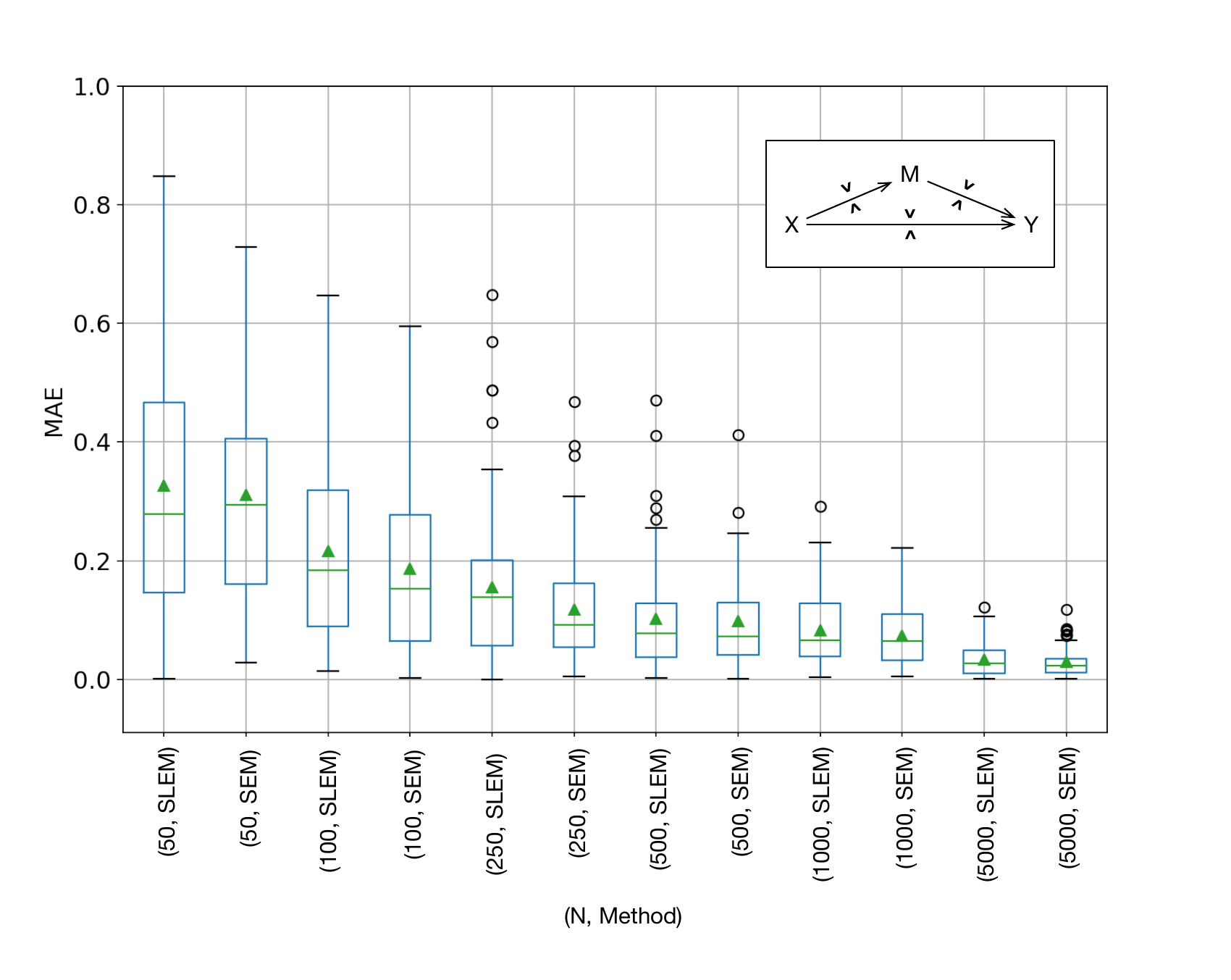}
\newline
\small{\textit{Note.} Mean Absolute Error (MAE) for the estimation of the true effect of the total effect of $X$ on $Y$. Green arrows represent the average MAE, whilst the horizontal lines depict the median.}
\label{fig:DGP_total}
\end{figure}

\begin{figure}[!ht]
\caption{Simple partial mediation SEM vs. the SLEM DAG Learner comparison simulation.} 
\includegraphics[width=0.8\linewidth]{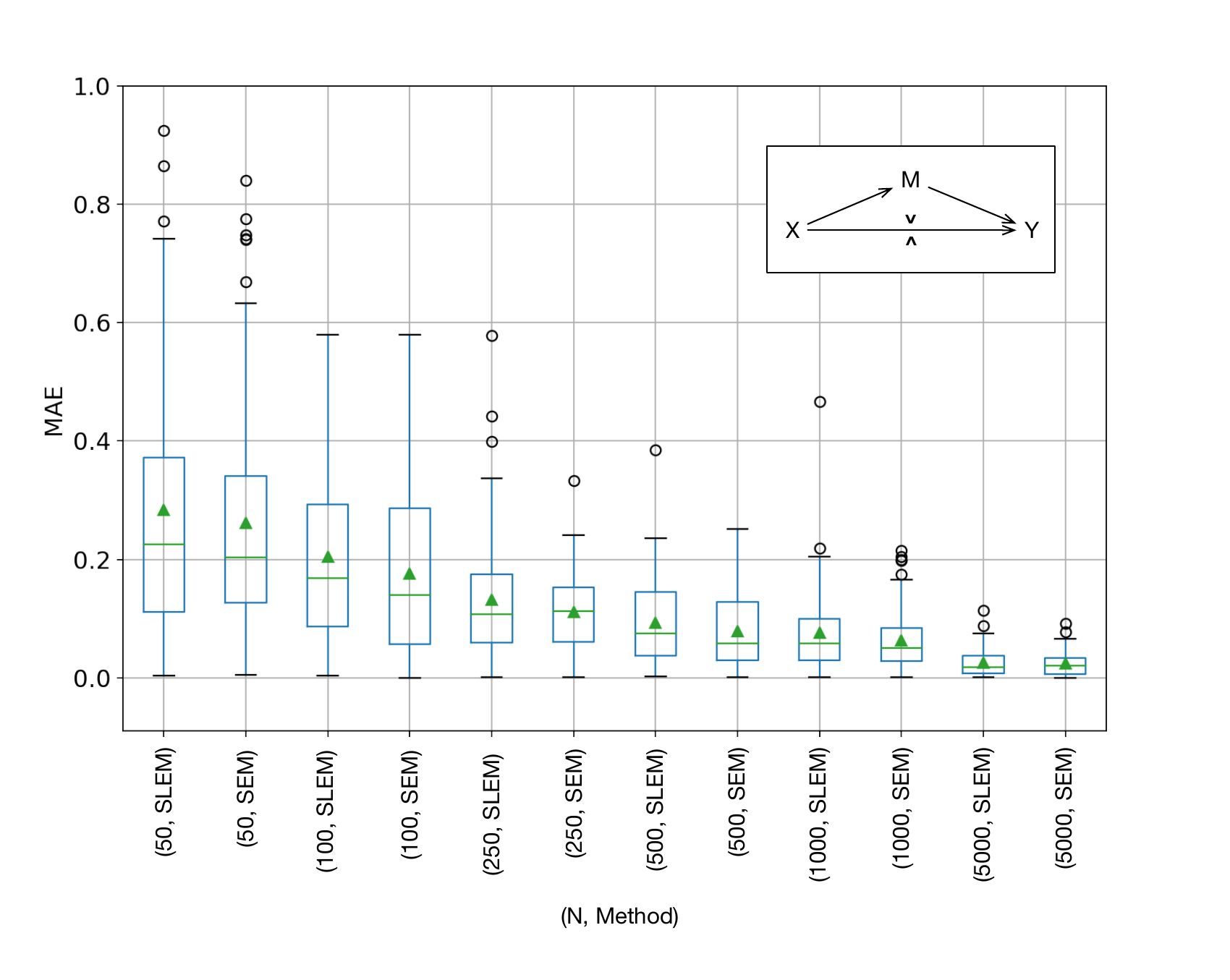}
\newline
\small{\textit{Note.} Mean Absolute Error (MAE) for the estimation of the true effect of the direct effect of $X$ on $Y$. Green arrows represent the average MAE, whilst the horizontal lines depict the median.}
\label{fig:DGP_direct}
\end{figure}

\begin{figure}[!ht]
\caption{Simple partial mediation SEM vs. the SLEM DAG Learner comparison simulation.} 
\includegraphics[width=0.8\linewidth]{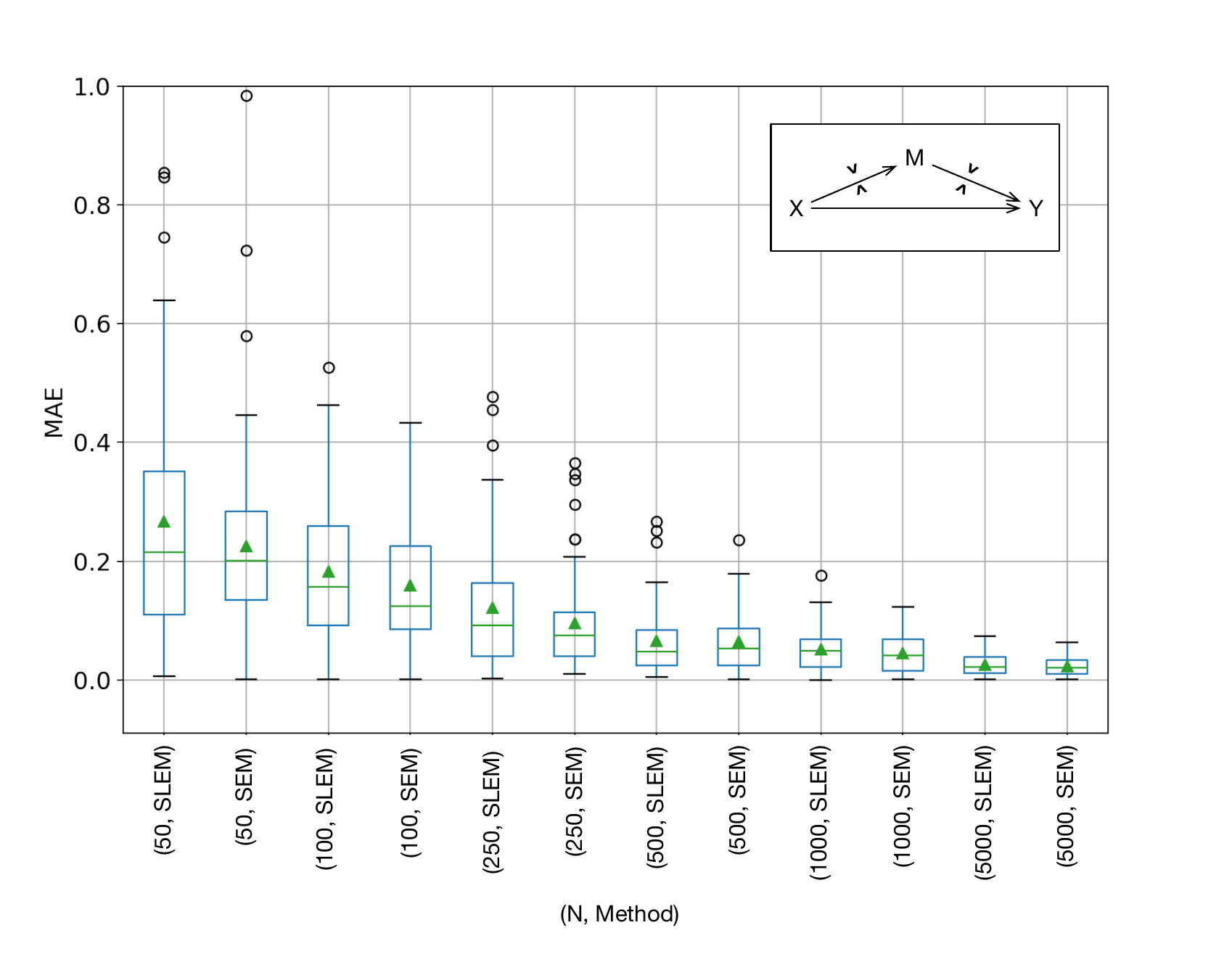}
\newline
\small{\textit{Note.} Mean Absolute Error (MAE) for the estimation of the true effect of the indirect effect of $X$ on $Y$. Green arrows represent the average MAE, whilst the horizontal lines depict the median.}
\label{fig:DGP_indirect}
\end{figure}

Two additional sets of simulation results are depicted in 
 Figures~\ref{fig:DGP_B}, \ref{fig:DGP_total}, \ref{fig:DGP_direct}, and \ref{fig:DGP_indirect}. These simulations undertaken by first generating 70 datasets according to the associated DGP/DAG and an associated set of SCM equations. This is then repeated for six different sample sizes: 50, 100, 250, 500, 1000, 5000, resulting in 420 datasets for each DGP. Finally, results are obtained for the mean absolute error of estimation for the effect of $X$ on $Y$ using the SLEM DAG Learner and compared against the effect size estimations obtained using a SEM \verb|lavaan| R library \parencite{Rosseel2012} specified (structurally) correctly according to the DAG. The DGP for Figure~\ref{fig:DGP_B} is:

 \begin{equation}
     \begin{split}
         U_{Z_{1-12}} \sim \mathcal{N}(0,1),  \\
         Z_{1-3} := U_{Z_{1-3}}, \; \; Z_{7,9,12}:= U_{Z_{7,9,12}}, \; \; Z_4 := 0.1Z_3 + U_{Z_4}, \; \; Z_5 := -0.2Z_2 + U_{Z_5},\\
         Z_6 := -0.3Z_5+ U_{Z_6}, \; \; Z_8 = 0.5Z_7 + U_{Z_8}, \; \; Z_{10} = 0.4Z_9 + U_{Z_{10}}, \; \;\\
         Z_{11} := 0.5Z_{10} + U_{Z_{11}}.
     \end{split}
 \end{equation}

 The DGP for Figures~\ref{fig:DGP_total}, \ref{fig:DGP_direct}, and \ref{fig:DGP_indirect} follows a linear, partial mediation structure:

 \begin{equation}
     \begin{split}
         U_M \sim \mathcal{N}(0,1), \; \; U_X \sim \mathcal{N}(0,1), \; \; U_Y \sim \mathcal{N}(0,1), \\
         X\sim Bi(p=\sigma(U_X)), \; \; M:= 0.8X + U_M, \; \;  Y := 0.5X +0.8M+U_Y
     \end{split}
 \end{equation}

In these equations,  $\sim$ indicates samples are randomly drawn from the corresponding distribution, $\mathcal{N(0,1)}$ denotes a standard normal distribution, $Bi(p)$ denotes a binomial distribution, and $\sigma$ denotes the sigmoid function.

 In terms of what can be seen from these simulations - both methods already provide unbiased estimation of the effect sizes (owing to the linearity of the DGP and the use of the DAG and \textit{do}-calculus to identify the causal effect) and both exhibit consistent estimation, demonstrated by the convergence of the MAE towards zero as sample size increases. The SLEM DAG Learner exhibits slightly worse average MAE than SEM in for both DGPs, although this difference decreases as sample size increases, and is unsurprising given that the SEM is actually perfectly specified both structurally and functionally.
 
\end{document}